\documentclass[letterpaper, 10 pt, conference]{ieeeconf}  
\usepackage{amsfonts}
\usepackage{amssymb}
\usepackage{amsmath}
\usepackage{graphicx} 
\usepackage{subfigure}
\usepackage{bm}
\usepackage{verbatim}
\usepackage{multirow}
\usepackage{makecell}
\usepackage{booktabs}
\usepackage{cite}
\usepackage{color}

\IEEEoverridecommandlockouts                              

\title{\LARGE \bf
REGNet: REgion-based Grasp Network for End-to-end Grasp Detection in Point Clouds
}

\author{Binglei Zhao, Hanbo Zhang, Xuguang Lan\textsuperscript{*}, Haoyu Wang, Zhiqiang Tian and Nanning Zheng 
\thanks{*Corresponding author: X. Lan. {\tt\small xglan@mail.xjtu.edu.cn}}
\thanks{*This work was supported in part by NSFC under grant No.62088102, No.91748208, NSFC No.61973246, Shaanxi Project under grant No.2018ZDCXLGY0607, and the program of the Ministry of Education.}
}

\begin{document}
\setlength{\abovecaptionskip}{0.cm}

\maketitle
\thispagestyle{empty}
\pagestyle{empty}

\begin{abstract}
Reliable robotic grasping in unstructured environments is a crucial but challenging task. 
The main problem is to generate the optimal grasp of novel objects from partial noisy observations.
This paper presents an end-to-end grasp detection network taking one single-view point cloud as input to tackle the problem. 
Our network includes three stages: Score Network (SN), Grasp Region Network (GRN), and Refine Network (RN). 
Specifically, SN regresses point grasp confidence and selects positive points with high confidence. 
Then GRN conducts grasp proposal prediction on the selected positive points. 
RN generates more accurate grasps by refining proposals predicted by GRN. 
To further improve the performance, we propose a grasp anchor mechanism, in which grasp anchors with assigned gripper orientations are introduced to generate grasp proposals. 
Experiments demonstrate that REGNet achieves a success rate of 79.34\% and a completion rate of 96\% in real-world clutter, which 
significantly outperforms several state-of-the-art point-cloud based methods, including GPD, PointNetGPD, and S$^4$G. 
The code is available at 
\emph{https:\//\//github.com\//zhaobinglei\//REGNet\_for\_3D\_Grasping}.\end{abstract}
\vspace{-0.04cm}

\section{Introduction}
\vspace{-0.1cm}
{Robotic grasping is a widely and actively investigated field in robotics since it plays a pivotal and fundamental role in manipulation and interaction with the outside world. 
However, reliable robotic grasping is still challenging due to the uncertainty caused by unstructured environments, sensor noise, and various object geometries.} Most grasp detection algorithms aim at generating stable grasps with high-quality scores, but the performance is far behind human beings.

Traditional methods\cite{c6,c7,c14,c13} based on physical analysis struggle to generate grasps of objects without 3D models. 
With a proliferation of deep-learning techniques, the superiority of data-driven methods for grasping unknown objects has been widely investigated\cite{survey1}.
Based on RGB images, Redmon et al. detect grasps represented by rectangles using CNNs\cite{c8}.
Nevertheless, due to lacking consideration of geometric information and 
{grasp quality metrics}, they struggle to find the optimal grasp and limit the way of grippers contacting objects (e.g., grippers usually perpendicular to the table).
On the other hand, some algorithms assess quality scores of sampled grasp candidates through trained classifiers\cite{c16, c17}. 
However, it is not easy to estimate normals accurately from noisy point clouds, causing the low efficiency of heuristic sampling methods based on the Darboux frame\cite{c18}.
Also, the execution time will go up quickly as the number of candidates increases. 
\begin{figure}[thpb]
\setlength{\abovecaptionskip}{0.cm}
\setlength{\belowcaptionskip}{-20.cm}
   \centering
   \includegraphics[height=3.1cm,width=8.1cm]{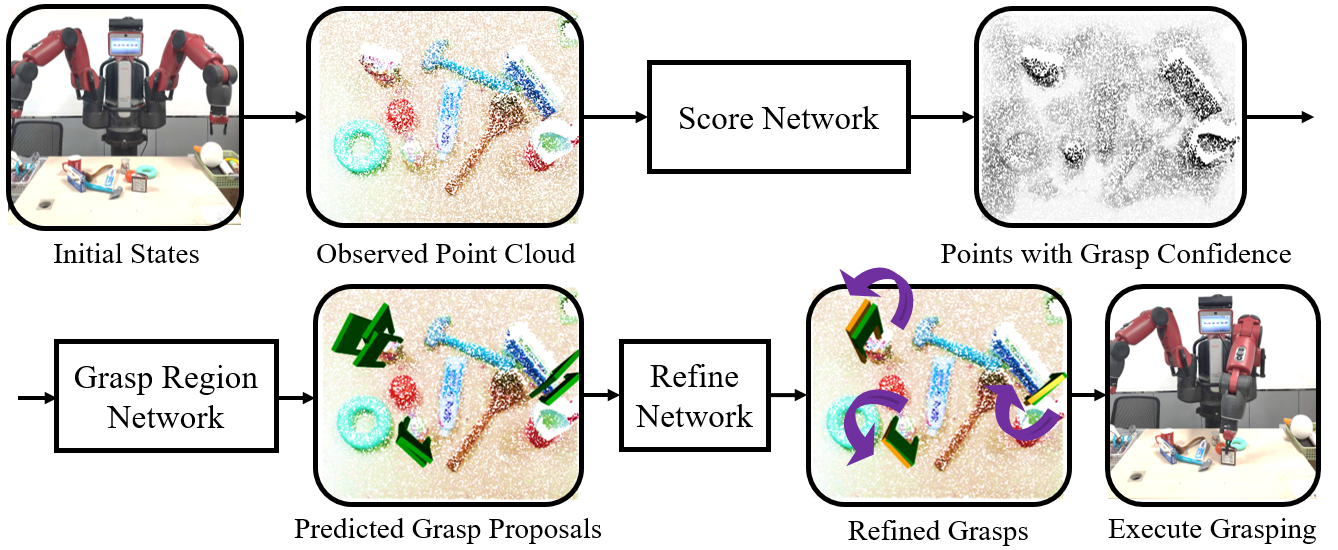}
   \vspace{-0.06cm}
   \caption{The pipeline. SN uses the observed point cloud as input and outputs points with grasp confidence. Then GRN predicts grasp proposals based on predicted confidence. After RN refines the proposals, we execute grasping.}
   \label{figurelabel}
   \vspace{-0.65cm}
\end{figure}
Recently, S$^4$G\cite{c18} directly regresses grasps using single-point features based on PointNet++\cite{c4} instead of sampling-then-evaluation. 
Since PointNet++ does not explicitly model the local spatial distribution of points, its single-point features contain less shape awareness\cite{c31},
leading to less accurate grasp detection.

In this paper, we present an end-to-end grasp detector to address the mentioned problems.
Our network consists of three stages: Score Network (SN), Grasp Region Network (GRN), and Refine Network (RN).
\textbf{SN} selects proper grasp positions in the whole scene.
\textbf{GRN} uses grasp region features instead of single-point features to predict grasp proposals, more effectively realizing local feature aggregation.
Compared to a single point, the grasp region contains surrounding points and provides a larger receptive field.
Besides, since anchor-based methods can achieve higher localization accuracy than anchor-free methods\cite{detection1}, we introduce the grasp anchor mechanism predefining grasp anchors with assigned orientations to more accurately predict proposals.
To further generate more accurate grasps, \textbf{RN} conducts proposal refinement by fully utilizing all information provided by SN and GRN.
Specifically, in the beginning, we use PointNet++\cite{c4} to extract features from single-view point clouds.
As shown in Fig. 1, SN predicts point grasp confidence and selects positive points with high confidence.
Afterwards, GRN constructs grasp regions centered on selected positive points to regress grasp proposals. 
Finally, RN refines the predicted proposals using fusion features of the grasp regions and the areas within the proposals.
To further train our proposed network, we create a dataset containing collision-free grasps with grasp quality scores and point grasp confidence in simulation.


To summarize, our contributions are as follows:
\begin{itemize}
\item We propose an end-to-end region-based grasp detection network to effectively predict grasps from partial noisy point clouds, which outperforms several state-of-the-art point-cloud based algorithms with remarkable margins, including GPD, PointNetGPD, and S$^4$G.
\item 
Our network can be successfully generalized to novel objects and transferred to the real-world clutter despite being trained only in simulation. It achieves a success rate of 79.34\% and a completion rate of 96\% in real robotic experiments.
\end{itemize}

\vspace{-0.2cm}
\section{Related Work}
\vspace{-0.15cm}
\subsection{Grasp Detection}
\vspace{-0.09cm}
Existing grasp detection methods are generally divided into two categories: model-based and model-free. 
{The model-based works obtain grasps on models using physical analysis\cite{c6,c7,c13,c14} and then register the observed points with models.
However, it is tricky to generalize them to generic objects without models,
while model-free methods based on deep learning can be conveniently transferred to novel objects.}

Many model-free methods first sample grasp candidates and then assess the quality scores. 
Lenz et al. present a sliding window approach to predict whether each patch contains a potential grasp\cite{c9}.
Dex-Net\cite{c11,dexnet4}, GPD\cite{c16} and PointNetGPD\cite{c17} tackle the problem by training classifiers based on depth maps and point clouds.
When input point clouds are sparse and noisy, the sampling method based on the Darboux frame has low efficiency\cite{c18}.
Mousavian et al. design a variational autoencoder to sample grasps rather than heuristic sampling\cite{c23,c24}.
Nevertheless, repetitive scanning candidates and classifying take lots of time.

In contrast,
Redmon et al. design an end-to-end network instead of sliding windows based on RGB-D images\cite{c8}.
Following it, some works\cite{c10, c25, c26, c27, c28, c29} make a series of improvements and achieve better performance.
GG-CNN\cite{ggcnn} predicts grasps for each pixel in depth maps.
However, they struggle to find the optimal grasp due to the lack of physics information and grasp quality scores.

With a proliferation of 3D vision-based deep-learning techniques, some works\cite{c18,c19,pointnet++grasp} focus on predicting grasps based on 3D feature extractors\cite{c4, c5, voxnet}.
The recent work, S$^4$G\cite{c18}, directly regresses grasps from single-point features extracted by PointNet++\cite{c4}.
{Though PointNet++ can extract the group information, single-point features acquire less shape awareness by lack of modeling the local spatial layout\cite{c31}, which leads to less accurate grasp detection.
So we present a network based on grasp regions and gripper closing areas to obtain local shape information of grasps effectively.}

\vspace{-0.2cm}
\subsection{Grasp Dataset} 
\vspace{-0.1cm}
The manually labeled datasets, such as Cornell\cite{cornell} and VMRD\cite{c25} grasp dataset, include no specific information about grasp quality metrics.
In contrast, some datasets are automatically generated through random trials or physic simulation\cite{jacquard, c30}. 
Some works also generate grasp datasets by analyzing the physical and geometric information.
{PointNetGPD\cite{c17} generates grasp quality scores in single-object scenes through grasp wrench space (GWS)\cite{c6,gws} and force-closure analysis \cite{c1, c2}, 
while S$^4$G\cite{c18} generates the scores in multi-object scenes.
To train REGNet, we construct not only collision-free grasps, but the point grasp confidence of each point in the observed point clouds, which defines the ratio of successful grasp in a region around a given point.
\begin{figure}[thbp]
\centering
\subfigure[]{
\begin{minipage}[t]{0.55\linewidth}
\centering
\includegraphics[height=2.4cm,width=3.8cm]{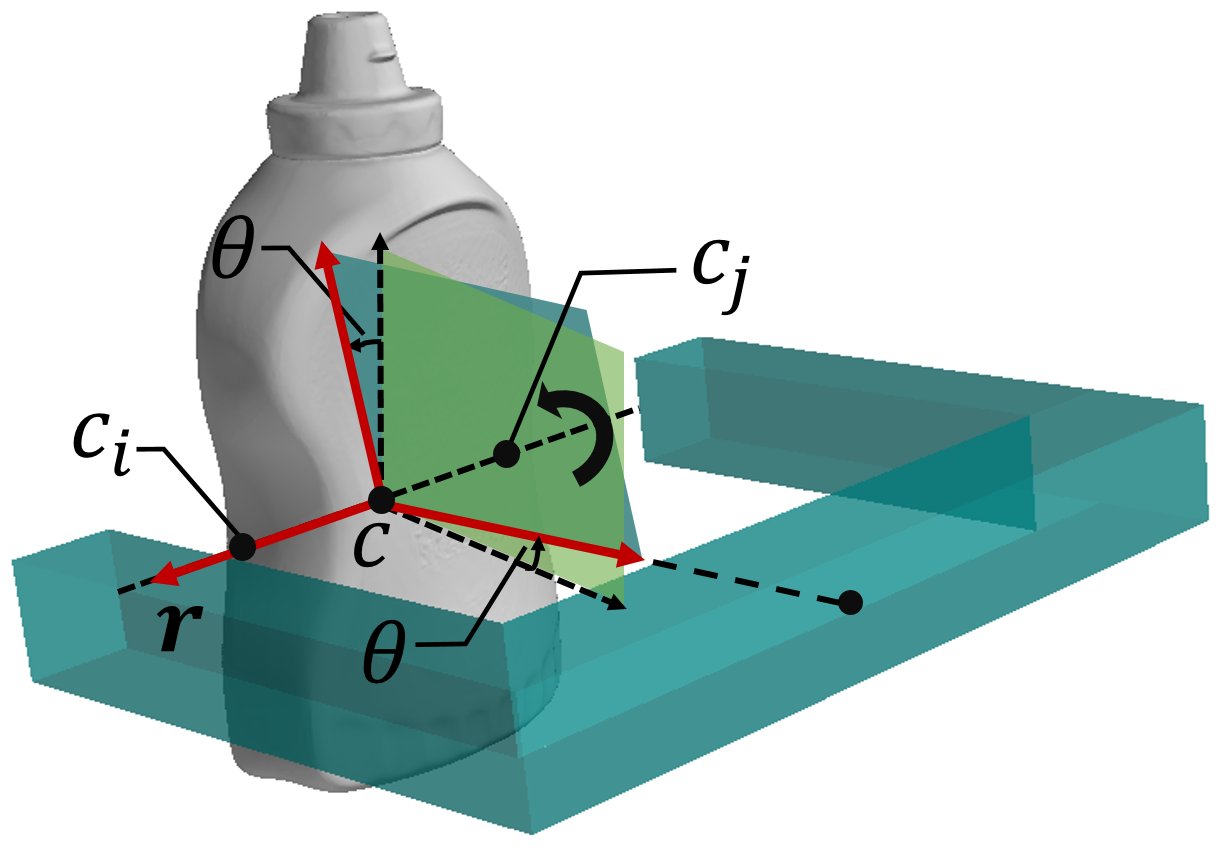}
\end{minipage}%
}%
\subfigure[]{
\begin{minipage}[t]{0.42\linewidth}
\centering
\includegraphics[height=2.4cm,width=2.25cm]{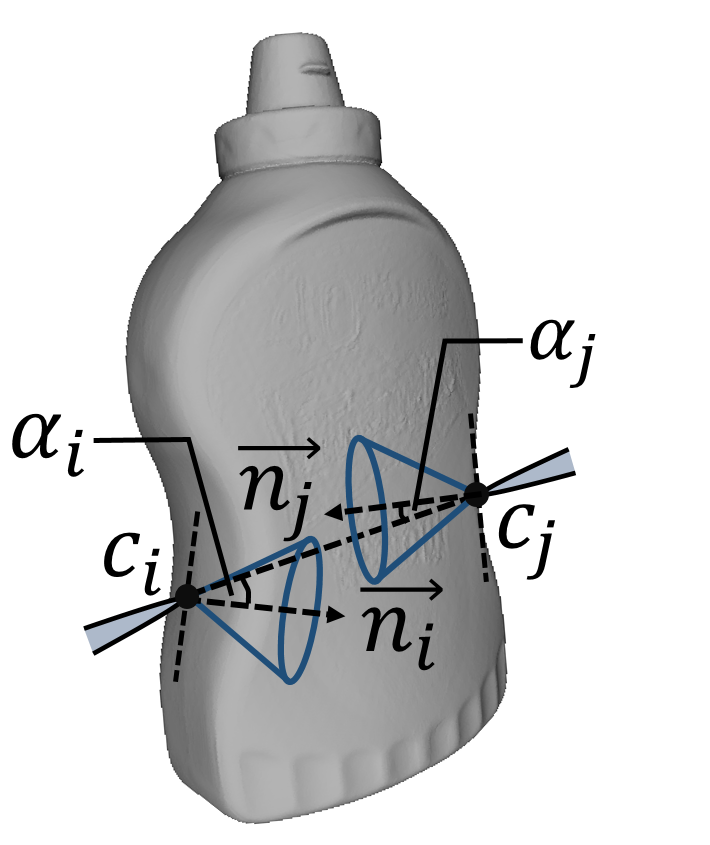}
\end{minipage}%
}%
\caption{(a) {The definition of a grasp}. 
(b) When a gripper closes to grasp, two contacts $c_i, c_j$ are generated.
$\alpha_i, \alpha_j$ are angles between the force direction and normals $n_i, n_j$. Blue lines show friction cones\cite{c1} generated at contacts.}
\vspace{-0.65cm}
\end{figure}

\vspace{-0.1cm}
\section{Problem Statement}
\vspace{-0.1cm}
Given an observed single-view point cloud $P$, we aim at learning parallel-jaw grasp configurations in 3D space. Based on PointNetGPD\cite{c17}, we define a \emph{grasp} as $g=(c,\bm{r},\theta)\in\mathbb{R}^7$, where $c=(x,y,z)\in \mathbb{R}^3, \bm{r}=(r_x,r_y,r_z)\in \mathbb{R}^3$ and $\theta\in [-{\pi/2},{\pi/2}]$ represent the grasp center, gripper orientation and approach angle, respectively, which is shown in Fig. 2(a). 
{Moreover, the corresponding \emph{grasp quality score} is defined as $s$, which means the grasp probability of a single grasp $g$.
We also define the \emph{point grasp confidence} $c_{p}$ as the ratio of successful grasp in a region around a given point $p \ (p\in P)$ to assess which position is suitable for grasping.}

\vspace{-0.07cm}
\section{Grasp Dataset Generation}
\vspace{-0.0cm}
{To train REGNet, we create a large-scale dataset $G$.
To recognize which position is proper for grasping, we define point grasp confidence as the ratio of successful grasp in a region around a given point.
When generating data, we simulate scenes and observed point clouds, 
construct collision-free grasps, and obtain grasp quality scores $s$ and point grasp confidence $c_{p}$.}

\textbf{Simulated scenes and point clouds generation.}
{Following S$^4$G\cite{c18}, we randomly select objects from YCB dataset\cite{c20} and load their models into MuJoCo\cite{mujoco}.
When they are equilibrium, we record their poses and positions.
The scene contains points of a table and objects with recorded poses and positions, while the point cloud contains the single-view noisy points observed by Kinect in {BlenSor}\cite{blensor}.}

\textbf{Collision-free grasps construction.} 
{If an object's density is constant, it is more stable to perform grasping near its center (barycenter).
Hence we sample more points as grasp centers closer to objects' centers in synthetic scenes.
Grasp candidates centered at the points are sampled in the Darboux frames' neighborhood.
Then we detect whether candidates collide with scenes and put collision-free grasps into $G$.}

\begin{figure*}[htpb]
   \centering
   \includegraphics[height=5.82cm,width=17.2cm]{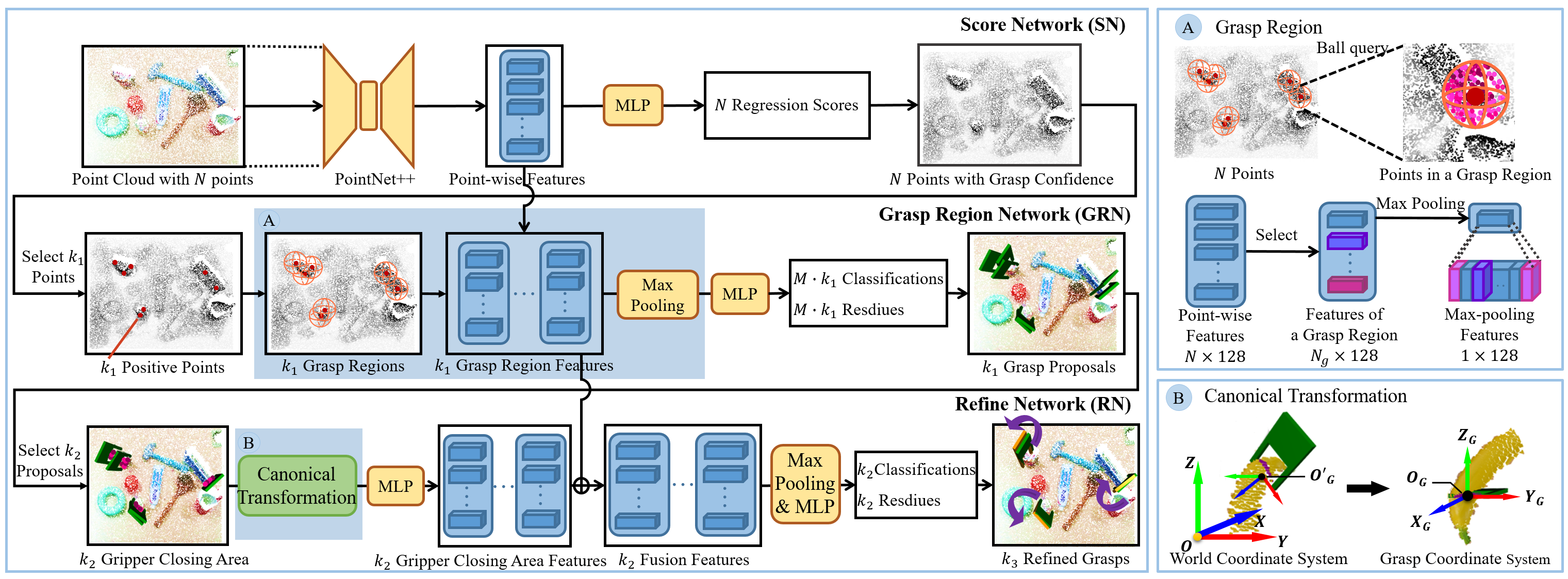}
   \vspace{-0.06cm}
   \caption{The architecture of REGNet. 
   SN takes the raw point cloud as input and outputs corresponding grasp confidence $c_p$. 
   The darker the color of the point in the point-grasp-confidence picture, the higher the confidence.
   GRN takes points with confidence generated by SN as input and predicts grasp proposals based on grasp region. 
   RN then refines proposals predicted by GRN and outputs final grasps with corresponding grasp quality score.
   (A) shows the grasp region and the grasp region features. 
   (B) demonstrates the point transformation from the world coordinate system to the grasp coordinate system.}
   \label{figurelabel}
   \vspace{-0.6cm}
\end{figure*}

\textbf{Grasp quality score generation.} 
{We define the grasp quality score $s$ as the antipodal score, describing the force closure property\cite{c2} of a grasp.
In Fig. 2(b), each grasp will generate two contact points $c_i,c_j$ with the object. 
Regardless of friction coefficients, 
the smaller the angles $\alpha_i,\alpha_j$ between the force direction (connection direction of two contacts) and contacts' normals, the greater the possibility of a grasp to be force-closure.
Similar to \cite{c18}, we define $s=cos\alpha_i \cdot cos\alpha_j$.
}

\textbf{Point grasp confidence generation.}
{
The point grasp confidence $c_{p}$ describes the ratio of successful grasp in a region around a given point $p$, which can help learn which position in $P$ is suitable for grasping.
Specifically, we count all grasps in $G$ to calculate $c_{p}$, which is defined as:}
$$ 
c_{p} = tanh\sum_{g \in G} \sigma_{g} , \ p \in P  \eqno{(1)}
$$
\vspace{-0.28cm}
$$
\sigma_{g}=
   \begin{cases}
   0&\mbox{$dis(p, c_g) \ \textgreater \ d_{th}$}\\
   1- dis(p, c_g) / d_{th}&\mbox{else}
   \end{cases}  
$$
where $g$ is a grasp in $G$, 
$dis(p, c_g)$ is the distance between $p$ and the center of ${g}$,
and $d_{th}$ is the distance threshold.
If $dis(p, c_g) \textgreater d_{th}$, $g$ does not contribute to the grasp confidence of point $p$.
Intuitively, $c_{p}$ will be higher as there are more grasp annotations near $p$.
In experiments, $d_{th}$ is set as $0.02m$.

\vspace{-0.1cm}
\section{Proposed Approach}
\vspace{-0.05cm}
We present the \textbf{RE}gion-based \textbf{G}rasp \textbf{N}etwork to detect grasps in 3D space. 
As illustrated in Fig. 3, the overall architecture includes three stages, Score Network (SN), Grasp Region Network (GRN), and Refine Network (RN). 

\vspace{-0.12cm}
\subsection{Score Network for Point Grasp Confidence Evaluation} 
\vspace{-0.1cm}
The \emph{Score Network} (SN) takes the observed point cloud $P$ as input to estimate \emph{point grasp confidence}, the ratio of successful grasp in a region around a given point.
We utilize PointNet++\cite{c4} as our backbone network to extract features of the point cloud. 
It encodes the input points into group features and decodes the group features into point-wise features through distance interpolation. 
Especially, the point-wise features are shared with all three stages.
Furthermore, we use a multilayer perceptron (MLP) to regress point grasp confidence from the extracted features.
The SN loss $L_1$ is defined based on the MSE loss, formulated as (2).
\vspace{-0.01cm}
$$
L_{1} = \frac{1}{N}\sum\limits_{p \in P} (c_p - \widehat{c}_p)^2 \eqno{(2)}
$$
where $N$ is the number of points in the point cloud $P$, 
$c_{p}$ and $\widehat{c}_{p}$ are the ground-truth and predicted grasp confidence of $p$.
The generation method of $c_{p}$ is described in Section \MakeUppercase{\romannumeral4}.
Given a threshold $c_{th}$, if $\widehat{c}_{p} > c_{th}$, $p$ is a suitable position for grasping.  
In the following, $P_{pos}= \left\{ p \ | \ \widehat{c}_{p} > c_{th}, p \in P \right\} $ is called the \emph{positive point set}.
In practice, we set $c_{th} = 0.5$.

\vspace{-0.1cm}
\subsection{Grasp Region Network for Grasp Proposal Generation}
\vspace{-0.05cm}
The \emph{positive point} in $P_{pos}$ has a high probability of generating suitable grasps, which holds valuable information for predicting associated grasp.
Instead of single-point features, \emph{Grasp Region Network} (GRN) uses \emph{grasp region features} to regress grasp proposals at positive points.
Compared to a single point, the \emph{grasp region} contains surrounding points and provides a larger receptive field, which effectively helps aggregate local features.
Furthermore, since anchor-based methods can achieve more accurate localization\cite{detection1}, we introduce the grasp anchor mechanism that predefines grasp anchors with assigned orientations to predict proposals.

Since most proposals centered on neighboring points are similar, it is unnecessary to use all points in $P_{pos}$ to regress. 
In $P_{pos}$, we only keep a subset containing $k_1$ points using the farthest point sampling method (FPS)\cite{c4}.
FPS ensures that our network can cover as many points with different locations and structures as possible.
Then we obtain $k_1$ grasp regions, spheres centered on the points in $P_{pos}$.
GRN obtains $k_1$ grasp region features and regresses the proposal on each of them, totally $k_1$ proposals, based on the grasp anchor mechanism.

\textbf{Grasp region. }
In the research of 2D object detection, ``Region" is often considered to be a rectangle\cite{c3}. Nevertheless, in this paper, \emph{Grasp Region} is a sphere centered on a \emph{positive point} ${p}_a \in P_{pos}$, which is illustrated in Fig. 3(A). 

We use ball query\cite{c4} to find all points within a radius $\phi$ to a positive point $p_a$, which guarantees the grasp region has a fixed region scale.
To ensure a fixed-dimensional input, we randomly sample and keep $N_g$ points in each grasp region.
In Fig. 3(A), we select extracted features of the $N_g$ points for each grasp region, which are called \emph{grasp region features} below.
Then max-pooling features are obtained from grasp region features through max-pooling operation.
Noticeably, since $P_{pos}$ has $k_1$ points, there are $k_1$ grasp regions.
By applying MLP on the $k_1$ max-pooling features, we obtain $k_1$ proposals based on the grasp anchor mechanism. 


\textbf{Grasp anchor mechanism. }
\begin{figure}[thpb]
   \centering
   \includegraphics[height=2.5cm,width=5cm]{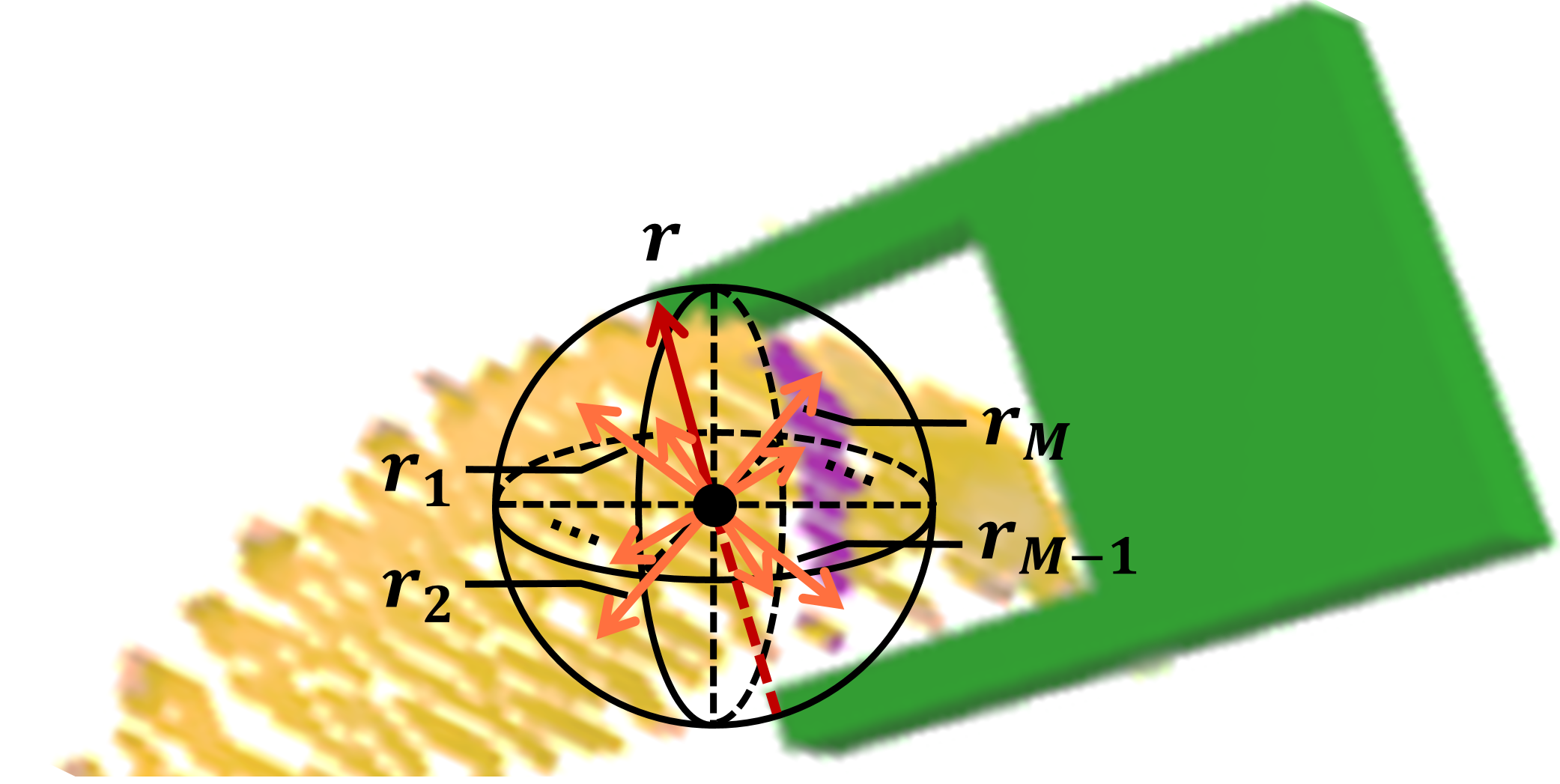}
   \caption{Illustration of predefined assigned orientations. The red line shows the ground-truth orientation $\bm{r}$.
   The orientation category is determined by the minimum angle between $\bm{r}$ and all assigned orientations (orange lines). }
   \label{figurelabel}
   \vspace{-0.5cm}
\end{figure}
Instead of direct regression, anchor-based methods can achieve higher localization accuracy\cite{detection1}.
{As mentioned, a grasp is defined as $g=(c,\bm{r},\theta)$.
Since small changes in $\theta$ are marginal for grasp prediction, $\theta$ is permitted to have an acceptable regression error. 
So we predefine the assigned approach angle to be $0$, which means the prediction of $\theta$ is simplified to direct regression.
We define the \emph{grasp anchor} as $g_a = ({p}_a, \bm{r_{i}}, \emph{0}),(i =1,2, ..., M)$, 
where $\bm{r_{i}}$ is the assigned orientation, and $M$ is the number of $\bm{r_{i}}$.
Each anchor at $p_a$ is associated with a 3-dimensional orientation $\bm{r_{i}}$ and a zero approach angle as the reference. 
As shown in Fig. 4, the assigned orientation $\bm{r_{i}}$ is sampled from the origin to a unit sphere surface. 
In practice, $M$ is set as $8$, and assigned orientations are $(\pm\sqrt{3}/{3},\pm{\sqrt{3}}/{3}, \pm{\sqrt{3}}/{3})$.}

The orientation estimation loss consists of two terms, one for classification of $\bm{r_{i}}$, and the other for residual regression. 
We directly utilize smooth L1 loss to estimate grasp centers because the distance between $p_a$ and its corresponding grasp center is within a small range. 
We also adopt direct regression to estimate the approach angle and the quality score.
The optimized targets at $p_a$ are as (3).
Specially, ${res}^{\theta}_{p_a}$ and ${res}^{s}_{p_a}$ are equal to ${\theta}_{p_a}$ and ${s}_{p_a}$, respectively.
$$
{o}_{p_a} = \mathop{\arg\min}_{i} \left\langle \bm{r_{i}}, \bm{r_{p_a}} \right\rangle , \  1 \leqslant i \leqslant M
$$
\vspace{-0.03cm}
$$
{res}^r_{p_a} =  \frac{\bm{r_{p_a}}}{\|\bm{r_{p_a}}\|} - \frac{\bm{r_{{{o}_{p_a}}}}}{\|\bm{r_{{{o}_{p_a}}}}\|}  \eqno{(3)}
$$
\vspace{-0.03cm}
$$
{res}^c_{p_a} = \left(c_{p_a} - p_a \right) / S
$$
where ${o}_{p_a}$ is the ground-truth orientation's category, 
${res}^u_{p_a}$ ($u \in \{ c, \bm{r}, \theta, s \}$) are the ground-truth residues of center, orientation within the category, approach angle and quality score, 
$u_{p_a}$ ($u \in \{ c, \bm{r}, \theta, s \}$) are the ground-truth grasp center, orientation, approach angle and quality score,
$\bm{r_{i}}$ and $\bm{r_{{{o}_{p_a}}}}$ are the $i$-th and the ${o}_{p_a}$-th assigned orientation,
and $S$ is the maximum of length, width and height of the gripper. 
During caculation, the unitization method guarantees $\bm{r_{p_a}}$ and $\bm{r_{{{o}_{p_a }}}}$ are unit vectors.
The overall GRN loss $L_2$ is defined as:
$$
L_2 = \frac{1}{N_{pos}} \left( \lambda_{cls} \cdot L_{cls} + \sum_{u \in \{c,\bm{r}, \theta , s\}} \lambda_u \cdot L_u \right) 
$$
\vspace{-0.03cm}
$$
L_{cls} =  \sum_{p_a \in P_{pos}} F_{cls} \left( {o}_{p_a},{\hat o}_{p_a} \right) \eqno{(4)}
$$
\vspace{-0.03cm}
$$
L_{u} = \sum_{p_a \in P_{pos}}  F_{reg}\left( {res}^u_{p_a}, \hat{res}^u_{p_a}\right) , \ u \in \{c,\bm{r}, \theta , s\}
$$
where $L_{cls}$ is the orientation's classification loss,
$L_u$ ($\ u \in \{c,\bm{r}, \theta , s\}$) are residual regression losses of center, orientation, angle and quality score. 
$N_{pos}$ is the number of points in $P_{pos}$, which is equal to $k_1$.  
${o}_{p_a}$ and ${\hat o}_{p_a}$ are the ground-truth and predited category of orientation,
${res}^u_{p_a}$ and $\hat{res}^u_{p_a}$ are the ground-truth and predited residues of center, orientation, angle and quality score.
$F_{cls}$ and $F_{reg}$ denote the cross-entropy classification loss and smooth L1 loss, respectively. 
Considering the different magnitudes of the losses, we set $\lambda_{cls}= 0.2$, $\lambda_{c}= 10$, $\lambda_{r}= 5$, $\lambda_{\theta}= 1$ and $\lambda_s= 1$ to balance the impact of each loss on $L_2$.
Finally, we predict $k_1$ grasp proposals based on the positive points.

\vspace{-0.09cm}
\subsection{Refine Network for Grasp Refinement}
\emph{Refine Network} (RN) fully utilize all information provided by SN and GRN to further obtain more accurate grasps.
The grasp proposal predicted by GRN is closer to ground truth.
Therefore, the area within the predicted grasp, called \emph{gripper closing area} contains information closer to the ground truth.
RN uses both the transformed gripper closing areas (features) and grasp region (features) for grasp proposal refinement.

Since an area containing fewer points has less information, we only use proposals containing more than $50$ points for refinement,
and the number of proposals for refinement is denoted as $k_2$.
We transform the points in $k_2$ gripper closing areas from the world coordinate system to grasp coordinate systems.
RN finally refines $k_2$ proposals using fusion features of gripper closing areas as well as grasp regions.

\textbf{Canonical transformation. }
For one predicted grasp $\hat{g} = ( \hat c, \hat{\bm r}, \hat{\theta})$, we transform each point in the gripper closing area from the world coordinate system to grasp coordinate system. 
As shown in Fig. 3(B), the \emph{grasp coordinate system} defines that: (1) the origin $O_G$ is located at $\hat c$; (2) $Y_G$ axis is along $\hat{\bm r}$; (3) $X_G$ axis is the approach direction, which is obtained by rotating $X'$ axis around $Y_G$ axis by $\theta$, ($X'$ is parallel to the ground in the world coordinate system and perpendicular to $Y_G$); (4) $Z_G$ axis is perpendicular to both $X_G$ and $Y_G$.

\begin{table*}[]
\centering
\setlength{\abovecaptionskip}{0.1cm}%
\setlength{\belowcaptionskip}{0cm}%
\caption{Results compared with baselines in simulation}
\setlength{\tabcolsep}{3.5mm}
\linespread{0.5}
\label{Tab01}
\begin{tabular}{lcccccccc}
\toprule[1pt]
\multirow{2}{*}{} & \multicolumn{2}{c}{Grasp Quality} & \multicolumn{3}{c}{Time Efficiency}\\
\cmidrule(r){2-3} \cmidrule(r){4-6} 
& Collision-free Ratio & Antipotal Score & Forward-passing Time & Processing Time & Total Time \\
\midrule
GPD (3 channels)  & 56.09\% & 0.3423 & \textbf{1.98ms} & 21128.69ms & 21130.67ms \\
GPD (12 channels) & 59.44\% & 0.3786 & 2.05ms & 23826.68ms & 23828.74ms \\
PointNetGPD       & 61.59\% & 0.3978 &{5.56ms} & 10685.55ms & 10691.11ms\\
S4G   & 72.84\% & 0.5268 & {22.23ms} & {824.20ms} & \textbf{846.43ms}\\
REGNet & \textbf{82.11\%} & \textbf{0.5690} & 224.02ms & \textbf{768.23ms} & 992.25ms\\ 
\bottomrule[1pt]
\end{tabular}
\vspace*{-0.55cm}
\end{table*} 
\textbf{Feature fusion for refinement. }
{
MLP is applied to extract local features of the transformed points in gripper closing areas, which are called \emph{gripper closing area features}.
RN concatenates the $k_2$ grasp region features and gripper closing area features to get fusion features that are further used to refine $k_2$ proposals through max-pooling layers and MLP.
This strategy fully utilizes the local shape information of grasps obtained from GRN.

{In RN, we only regress residues of predicted proposals that are close to the ground truth.
We introduce $y$ to measure the similarity between the predicted proposal and the ground truth.
If the differences between the ground truth and the predicted values of orientation and approach angle are less than $2 \pi/9$ and $\pi / 3$, $y$ will be set as $1$. If else, $y = 0$. For each proposal with $y=1$, the residual targets are as follows.}
$$
{res}^{Rc} = \left( c - \hat{c} \right) / S ,\ \hat{c} = \hat{res}^c \cdot S + p_a   
$$
\vspace{-0.1cm}
$$
{res}^{Rr} =  \frac{\bm{r}}{\|\bm{r}\|} - \frac{\bm{\hat{r}}}{\|\bm{\hat{r}}\|} ,\ \bm{\hat{r}} = \hat{res}^r + \bm{r_{\hat{o}}} \eqno{(5)}
$$
\vspace{-0.1cm}
$$
{res}^{Ru} =  u - \hat{res}^{u}, \ (u \in \{ \theta, s \})
$$
where ${res}^{Ru}$ ($u \in \{ c, \bm{r}, \theta, s \}$) are the ground-truth residues of center, orientation, angle and quality score in RN,
$\hat{res}^u$ are the residues predited from GRN, 
$\hat{o}$ is the predicted gripper orientation's category,
$c, \bm{r}, \theta, s$ are the ground truth,
and $p_a$ is the center of the corresponding grasp anchor. 
Since the predicted proposals are close to the ground truth, RN directly regresses residues. The RN loss $L_3$ is defined as:
$$
L_3 = \frac{1}{k_2} \cdot \lambda'_{cls} \cdot L'_{cls} + \frac{1}{k_3} \cdot \sum_{u \in \{c,\bm{r}, \theta , s\}} \lambda'_u \cdot L'_u  
$$
\vspace{-0.05cm}
$$
L'_{cls} = \sum_{i=0}^{\substack{k_2}} F_{cls}\left( {y}_i, \hat{y}_i \right), \ y_i \in \{0, 1\}  \eqno{(6)}
$$
\vspace{-0.05cm}
$$
L'_{u} = \sum_{\substack{i=0\\y_i=1}}^{\substack{k_2}}  F_{reg}\left( {res}^{Ru}_{i}, \hat{res}^{Ru}_{i}\right)  , \ u \in \{c,\bm{r}, \theta , s\}
$$
where $L'_{cls}, \ L'_u$ are losses of classification and regressions,
$y_i$, ${res}^{Ru}_{i}$ are the ground-truth category and residues of the $i$-th grasp proposal, 
$\hat{y}_i$, $\hat{res}^{Ru}_{i}$ are the predicted category and residues,
and $k_3$ is the number of proposals with ${y}_i=1$.
Regression losses are only computed for grasps with ${y}_i=1$. 
$F_{cls}$ and $F_{reg}$ also denote the cross-entropy loss and smooth L1 loss.
In practice, we set $\lambda'_{cls}$ and $\lambda'_{u}$ all to $1$.

\vspace{-0.2cm}
\subsection{Implementation Details}
\vspace{-0.09cm}

{In the grasp dataset, each scene contains about 5-20 objects. For each object, 300 points are sampled as the centers of grasp candidates.
We generate 4794 clutter scenes using 131 object models and render 19176 observed noisy point clouds from 4 different perspectives.
We split the dataset by a ratio of $4:1$ for training and test.}

Moreover, the network details are as follows.
SN randomly samples $N=25600$ points from the raw point cloud as input.
The points have $6$ dimensions, including location (x, y, z) and color (r, g, b) information. 
In GRN, we set $k_1=64$ and $N_g=256$ during training. Noticeably, $k_1$ can be changed during test.
$\phi$ is the half maximum of the gripper's length, width, and height. 
All stages are trained simultaneously with the batch size $6$ and learning rate $0.001$ in the beginning. 
We divide the learning rate by $2$ every $5$ epochs. 

\section{Experiments}
We evaluate REGNet both in simulation and the real world.
REGNet significantly outperforms several point-cloud based methods in terms of collision-free ratio and antipodal score.
Tested on a new synthetic dataset with different distributions, REGNet can be successfully generalized to novel objects and unstructured environments.
The ablation studies illustrate that each component of REGNet has an inextricable effect on the performance.
Moreover, our method achieves a success rate of 79.34\% and a completion rate of 96\% for grasping novel objects using parallel-jaw grippers in real robotic experiments.
\begin{table}
\vspace{0.15cm}
\begin{center}
\setlength{\abovecaptionskip}{0.2cm}%
\setlength{\belowcaptionskip}{0cm}%
\setlength{\tabcolsep}{0.81mm}
\caption{Comparison results on novel objects in Simulation}
\begin{tabular}[l]{@{}llccccc}
\toprule[1pt]
 &&Simple&Semi-dense&Dense&Stacked&Total\\
\midrule
\multirow{2}*{S$^4$G}&Collision-free & 64.85\% & 62.84\% & 63.64\% & 63.66\% & 63.75\%\\
~ & Antipotal Score & 0.4873 & 0.4729 & 0.4658 & 0.4504 & 0.4691\\
\midrule
\multirow{2}*{REGNet}&Collision-free & 71.99\% & 72.11\% & 72.36\% & 72.23\% & 72.17\%\\
~ & Antipotal Score & 0.4755 & 0.4734 & 0.4582 & 0.4719 & 0.4698\\
\bottomrule[1pt]
\end{tabular}
\end{center}
\vspace{-0.7cm}
\end{table}

\vspace{-0.09cm}
\subsection{Simulation Experiments}
\vspace{-0.09cm}
\textbf{Baselines and Evaluation Metrics.}
The compared methods include 3-channel and 12-channel versions of GPD\cite{c16}, 2-class single-view PointNetGPD\cite{c17}, as well as S$^4$G\cite{c18}.
In experiments of GPD and PointNetGPD,
we firstly sample grasps that do not collide with the observed point cloud and then evaluate their quality scores using the trained classifier.
We all use a score threshold $s_{th} = 0.5$ to filter the grasps with higher predicted quality scores for a fair comparison.

Following S$^4$G\cite{c18}, we use two metrics to evaluate the quality of generated grasps: collision-free ratio and antipodal score, which describe the possibility of not colliding with the entire scene and force closure property of grasps, respectively.
We also compare the forward-passing time and processing time, including pre-processing and post-processing.

\textbf{Results and analysis.}
Results of simulation experiments demonstrate that REGNet significantly outperforms all baseline methods and is successfully generalized to novel objects.

In Table \MakeUppercase{\romannumeral1}, 
our algorithm increases the {collision-free ratio by 9.27\% and antipodal score by 0.0422} compared with S$^4$G, which indicates its efficiency for grasp detection in 3D space.
It also shows the advantages of end-to-end algorithms compared to algorithms that sample first and then evaluate, such as GPD and PointNetGPD.
Later ablation experiments will analyze the reason why REGNet outperforms S$^4$G.
As for the time, REGNet is comparable to the fastest algorithm, S$^4$G.
Its complex network structure increases 
the forward-passing time, while the end-to-end structure reduces the processing time after network forward propagation. 
Moreover, we can choose alternative feature extraction networks\cite{c31, c32} as the backbone to reduce the forward-passing time.

\begin{table}
\begin{center}
\setlength{\abovecaptionskip}{0.2cm}%
\setlength{\belowcaptionskip}{0cm}%
\setlength{\tabcolsep}{0.92mm}
\renewcommand\arraystretch{0.9} 
\caption{Results of ablation anaysis}
\begin{tabular}[l]{@{}lccccc}
\toprule[1pt]
 &\makecell[c]{Grasp \\ Region}&\makecell[c]{Grasp \\Anchor} &\makecell[c]{RN\\ Stage}&\makecell[c]{Collision-free \\Ratio}&\makecell[c]{Antipotal \\Score}\\
\midrule
\ \ \ \ S$^4$G & $\times$ & $\times$ & $\times$ &72.84\% & 0.5268 \\
\ \ \ \ Ours(direct-single) & $\times$ & $\times$ & $\times$ &{74.04\%} & {0.4922} \\ 
\ \ \ \ Ours(direct-region) & \checkmark & $\times$ & $\times$ & {77.34\%} & {0.5028} \\ 
\ \ \ \ Ours(w/o RN) & \checkmark & \checkmark & $\times$ & 80.15\% & 0.5346 \\ 
\ \ \ \ Ours\_xyz(w/ \ RN) & \checkmark & \checkmark & \checkmark & 81.57\% & 0.5662 \\
\ \ \ \ Ours(w/ \ RN) & \checkmark & \checkmark & \checkmark & \textbf{82.11\%} & \textbf{0.5690} \\
\bottomrule[1pt]
\end{tabular}
\end{center}
\vspace{-0.4cm}
\end{table}
\begin{figure}[htpb]
   \centering
   \includegraphics[height=3.2cm,width=7.4cm]{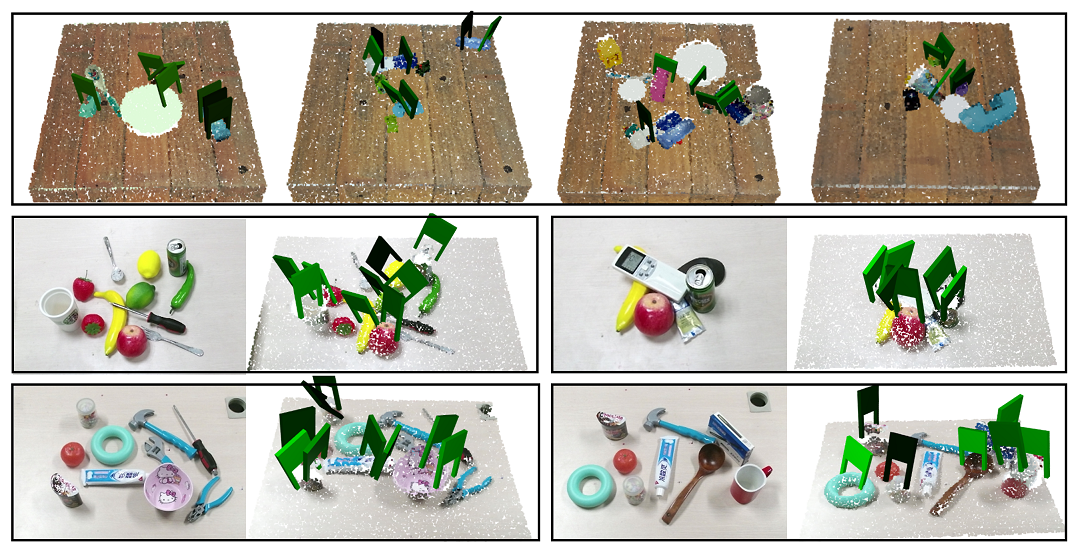}
   \caption{Predicted grasps on novel objects and in the real world. The darker the color of the grasp, the higher the corresponding grasp quality score.}
   \label{figurelabel}
   \vspace{-0.6cm}
\end{figure}

To evaluate the effectiveness of REGNet on novel objects and in unstructured environments, we construct a new dataset with different distributions using 38 objects not in the training dataset.
The new dataset contains scenes with different object densities, including
simple (1-5 objects), semi-dense (6-10 objects), dense (11-15 objects), and stacked (6-10 objects) scene types.
It has 600 test scenes and 2400 point clouds, where each type includes 150 scenes and 600 point clouds observed from 4 perspectives.
As shown in Table \MakeUppercase{\romannumeral2}, despite some performance degradation, REGNet can still effectively predict collision-free grasps on novel objects in different scenes, even in stacked scenarios.
The first row of Fig. 5 shows grasps predicted by REGNet in simple, semi-dense, dense, and stacked scenes from left to right.

\vspace{-0.1cm}
\subsection{Ablation Analysis and Discussion}
\vspace{-0.09cm}
We conduct a series of ablation experiments. The results demonstrate that each principal component of REGNet contributes inextricably to the final performance.

\textbf{Evaluation of the Grasp Region.} 
For a fair comparison, we design the \emph{direct-single} and \emph{direct-region} versions of REGNet, which directly regress grasps from single-point and grasp region features, respectively.
They both include the SN stage, and the only difference is the feature used in GRN.
The collision-free ratio and antipodal score drop without the grasp region, manifesting the effectiveness of the grasp region.

\textbf{Evaluation of Grasp Anchor Mechanism.}
We design the \emph{direct-region} and \emph{w/o RN} versions for comparision.
The \emph{direct-region} version directly regresses grasps while the \emph{w/o RN} version regresses based on the grasp anchor mechanism, whose difference is whether using the mechanism.
Table \MakeUppercase{\romannumeral3} demonstrates that the grasp anchor mechanism contributes inextricably to the final performance.

\textbf{Evaluation of Refine Network.}
To analyze RN's efficiency, we compare the \emph{w/o RN} and \emph{w/ RN} versions, whose difference is containing RN or not.
Removing RN decreases the collision-free ratio by 1.96\%, which shows that 
RN facilitates improvements in grasp detection capacity.

\textbf{Evaluation of Using RGB Features.}
To analyze the effort of the RGB features of point clouds, we design the \emph{ours\_xyz} version using only XYZ features. Except for the number of input channels, it has the same network architecture with \emph{w/ RN} version. As illustrated in Table \MakeUppercase{\romannumeral3}, using RGB features increases both the collision-free rate and antipodal score.

Noticeable, S$^4$G and our direct-single version both directly regress grasps using single-point features.
\begin{table}[htpb]
\begin{center}
\setlength{\abovecaptionskip}{0.2cm}%
\setlength{\belowcaptionskip}{0cm}%
\setlength{\tabcolsep}{4.9mm}
\renewcommand\arraystretch{0.93} 
\caption{Results of robotic experiments}
\begin{tabular}[l]{@{}lcc}
\toprule[1pt]
 &Success Rate & Completion Rate\\
\midrule
\ \ \ \ GPD (3 channels) & 45.87\% & 50.00\% \\
\ \ \ \ GPD (12 channels) & 46.23\% & 49.00\% \\
\ \ \ \ PointNetGPD & 49.06\% & 52.00\% \\
\ \ \ \ S$^4$G & 76.67\% & 92.00\% \\
\ \ \ \ REGNet & \textbf{79.34}\% & \textbf{96.00\%} \\
\bottomrule[1pt]
\end{tabular}
\end{center}
\vspace{-0.3cm}
\begin{flushleft}
\scriptsize{The success rate is the ratio of successful grasps, and the completion rate is the ratio of objects removed from the clutter.
}\\
\end{flushleft}
\vspace{-0.48cm}
\end{table}
\begin{figure}[htpb]
   \centering
   \includegraphics[height=1.9cm,width=7cm]{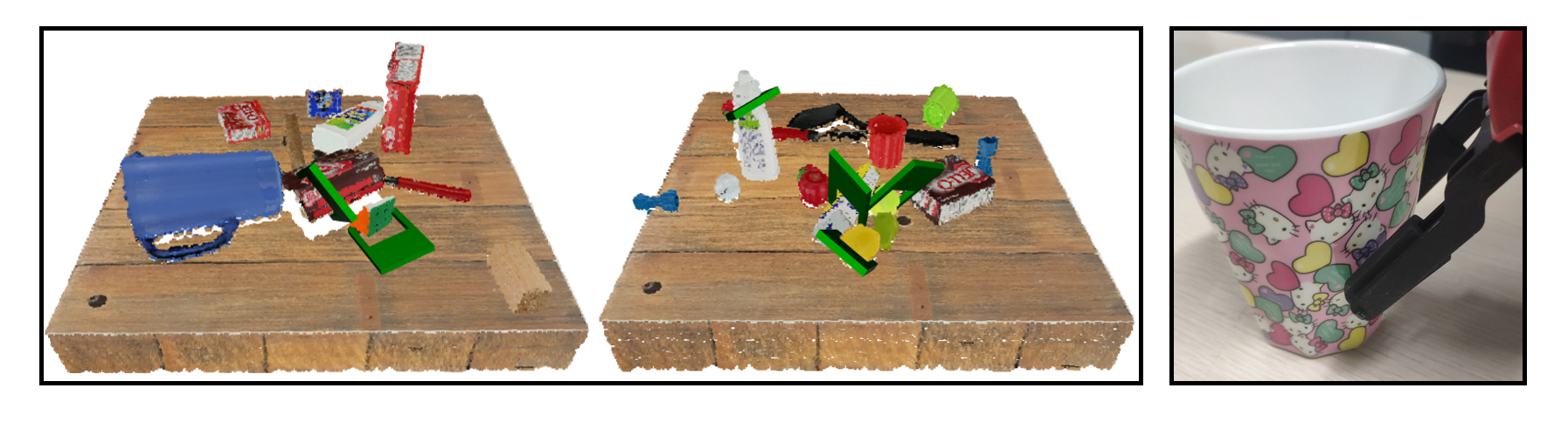}
	\vspace{-0.2cm}
   \caption{Left: some side grasps. Right: some failure cases.}
   \label{figurelabel}
\vspace{-0.65cm}
\end{figure}
Since S$^4$G utilizes a continuous representation of rotation matrix, it can generate grasps with higher antipodal scores. 
The comparison results of REGNet and S$^4$G show the effectiveness of GRN and RN.

\vspace{-0.05cm}
\subsection{Real Robotic Experiments}
\vspace{-0.08cm}
The robotic experiments are tested on Baxter with 2-finger parallel-jaw grippers (Fig. 1). We use the KinectV2 mounted on the head of Baxter to observe point clouds. 
Similar to the experimental procedure of S$^4$G\cite{c18}, we randomly choose ten objects to compose a cluttered scene. 
The robot performs grasping in each scene until no grasp is detected, or 15 trials are performed.
Each method in Table \MakeUppercase{\romannumeral4} is tested in 10 scenes. 
Same as S$^4$G, we adopt the success rate and completion rate as evaluation metrics.
It is noteworthy that all objects in robotic experiments are not in the training dataset.

The last two rows of Fig. 5 show captured point clouds and some grasp detection results using REGNet in real-world scenarios.
In Table \MakeUppercase{\romannumeral4}, our method outperforms all benchmarks, which suggests that REGNet can effectively generate stable grasps with high-quality scores despite being trained only on synthetic data.
In GPD and PointNetGPD, grasp candidates are sampled in the neighborhood of the Darboux frame.
Nevertheless, it is challenging to estimate the Darboux frame from observed noisy point clouds accurately, resulting in low efficiency of these algorithms.
As shown in Fig.6, REGNet can generate side grasps to deal with the situation where top-down grasps cannot be performed.
However, there are some failure cases due to the uncertainty of objects' friction coefficients and the difficulty of selecting a proper score threshold in the real world.
When grasping a slippery object, like a cup, the grasp with a high antipodal score may not be force-closure, causing a failure of grasping.

\vspace{-0.05cm}
\section{CONCLUSIONS}
\vspace{-0.06cm}
We present the REGNet, an end-to-end region-based network based on the single-view partial point cloud to detect grasps in 3D space. 
It contains three stages: SN, GRN, and RN, which significantly improve the grasp detection performance in dense clutter.
Moreover, REGNet is only trained on synthetic data and successfully transfers to detect grasps on novel objects in real-world scenarios. 

In future work, we will improve our network structure for target-driven object grasping in clutter. 
Furthermore, we will also detect objects' relationships to generate an appropriate grasping sequence in stacked scenes.

\addtolength{\textheight}{0cm}   






\end{document}